%% file: main.tex
\documentclass[10pt,twocolumn,letterpaper]{article}

\usepackage[pagenumbers]{cvpr} 

\usepackage{booktabs}       
\usepackage{amsfonts}       
\usepackage{graphicx}
\usepackage{amsmath}
\usepackage{pifont}
\usepackage{algorithm}
\usepackage[noend]{algpseudocode}
\usepackage{xcolor}         
\usepackage{color, colortbl}
\usepackage{multirow}
\usepackage{tabularx}

\usepackage[pagebackref,breaklinks=true,colorlinks,citecolor=blue,urlcolor=blue,linkcolor=blue,bookmarks=false]{hyperref}

\usepackage[capitalize]{cleveref}
\crefname{section}{Sec.}{Secs.}
\Crefname{section}{Section}{Sections}
\Crefname{table}{Table}{Tables}
\crefname{table}{Tab.}{Tabs.}

\def\cvprPaperID{*****} 
\def\confName{CVPR}
\def\confYear{2022}

\begin{document}


\title{DST-Det: Open-Vocabulary Object Detection via Dynamic Self-Training}

\author{Shilin Xu$^{1}$\thanks{The first two authors contributed equally to this work. \textsuperscript{$\dagger$} Project Leader.}, Xiangtai Li$^{2,3*} \textsuperscript{$\dagger$}$, Size Wu$^2$, Wenwei Zhang$^{2,3}$,  Yunhai Tong$^1$, Chen Change Loy$^2$ \\
  \small{$^{1}$Peking University}
  \small{$^{2}$S-Lab, Nanyang Technological University} 
   \small{$^{3}$Shanghai AI Laboratory} \\
 \small{Code: \url{https://github.com/xushilin1/dst-det} }
}

\maketitle

\input{latex/0_abs}

\input{latex/1_intro}

\input{latex/2_related_work}

\input{latex/3_method}

\input{latex/4_exp}

\input{latex/5_conclusion}

{\small
\bibliographystyle{ieee_fullname}
\bibliography{egbib}
}

\end{document}

%% file: latex/0_abs.tex
\begin{abstract}
Open-vocabulary object detection (OVOD) aims to detect the objects \textit{beyond} the set of classes observed during training.
This work introduces a straightforward and efficient strategy that utilizes pre-trained vision-language models (VLM), like CLIP, to identify potential novel classes through zero-shot classification. 
Previous methods use a class-agnostic region proposal network to detect object proposals and consider the proposals that do not match the ground truth as background. 
Unlike these methods, our method will select a subset of proposals that will be considered as background during the training.
Then, we treat them as novel classes during training. 
We refer to this approach as the self-training strategy, which enhances recall and accuracy for novel classes without requiring extra annotations, datasets, and re-training.
Compared to previous pseudo methods, our approach does not require re-training and offline labeling processing, which is more efficient and effective in one-shot training.
Empirical evaluations on three datasets, including LVIS, V3Det, and COCO, demonstrate significant improvements over the baseline performance without incurring additional parameters or computational costs during inference. 
In addition, we also apply our method to various baselines.
In particular, compared with the previous method, F-VLM, our method achieves a 1.7\% improvement on the LVIS dataset. 
Combined with the recent method CLIPSelf, our method also achieves 46.7 novel class AP on COCO without introducing extra data for pertaining. 
We also achieve over 6.5\% improvement over the F-VLM baseline in the recent challenging V3Det dataset.
We release our code and models at \url{https://github.com/xushilin1/dst-det}.
\end{abstract}

%% file: latex/1_intro.tex
\section{Introduction}
\label{sec:intro}

\begin{figure}[ht!]
    \centering
    \includegraphics[width=0.50\textwidth]{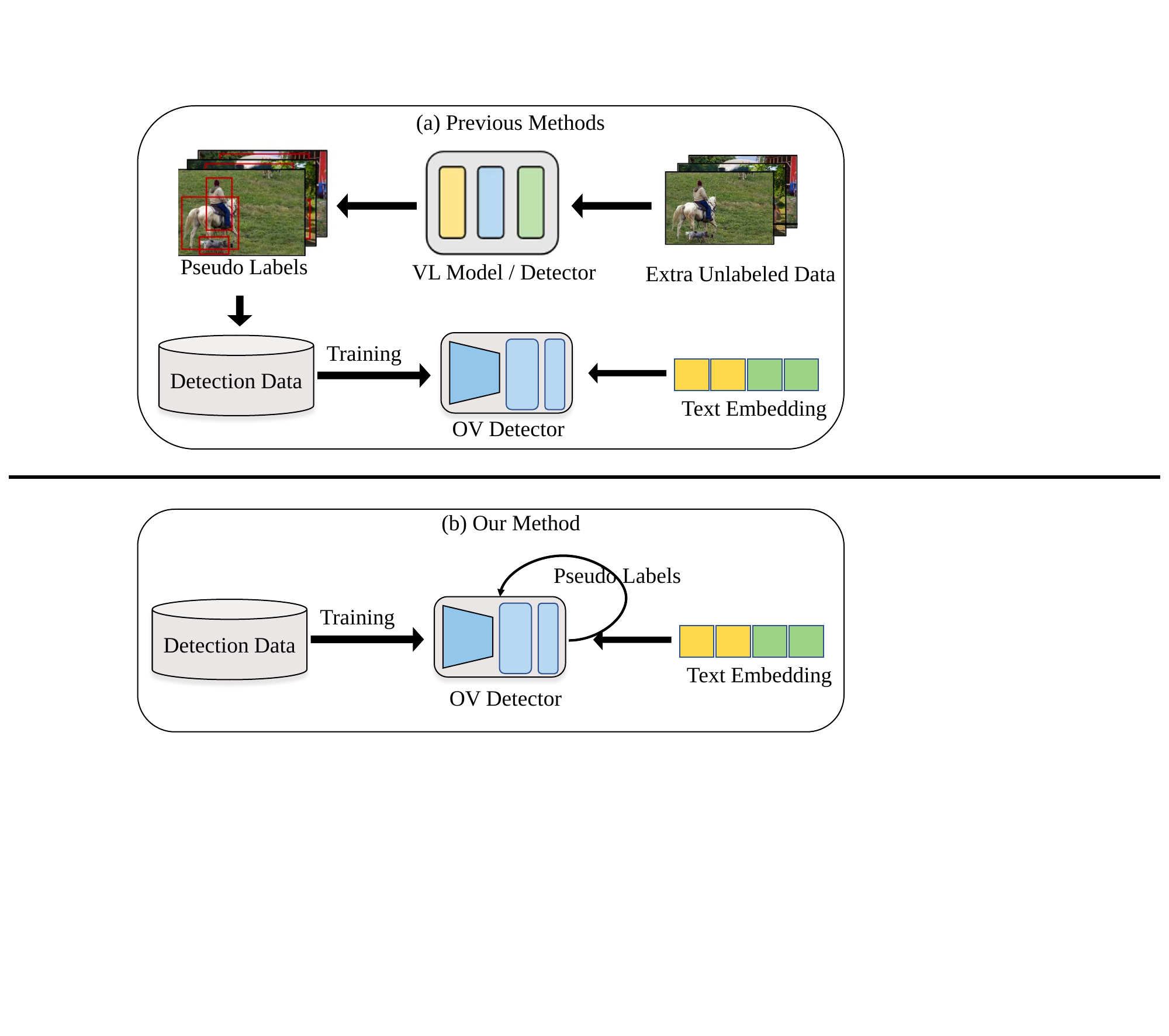}
    \caption{Compared with previous methods using the pseudo labels. \textbf{(a).} Previous methods~\cite{pb-ovd, vl-plm, zhao2023improving} usually obtain pseudo labels from extra unlabeled data. They first train a class-agnostic detector to generate object proposals and classify them using VLMs. \textbf{(b).}  Our DST-Det constructs an end-to-end pipeline, generates pseudo labels during training, and does not use  any extra data.}
    \label{fig:teaser2}
\end{figure}

\begin{figure*}[h]
\centering
\includegraphics[width=1.0\textwidth]{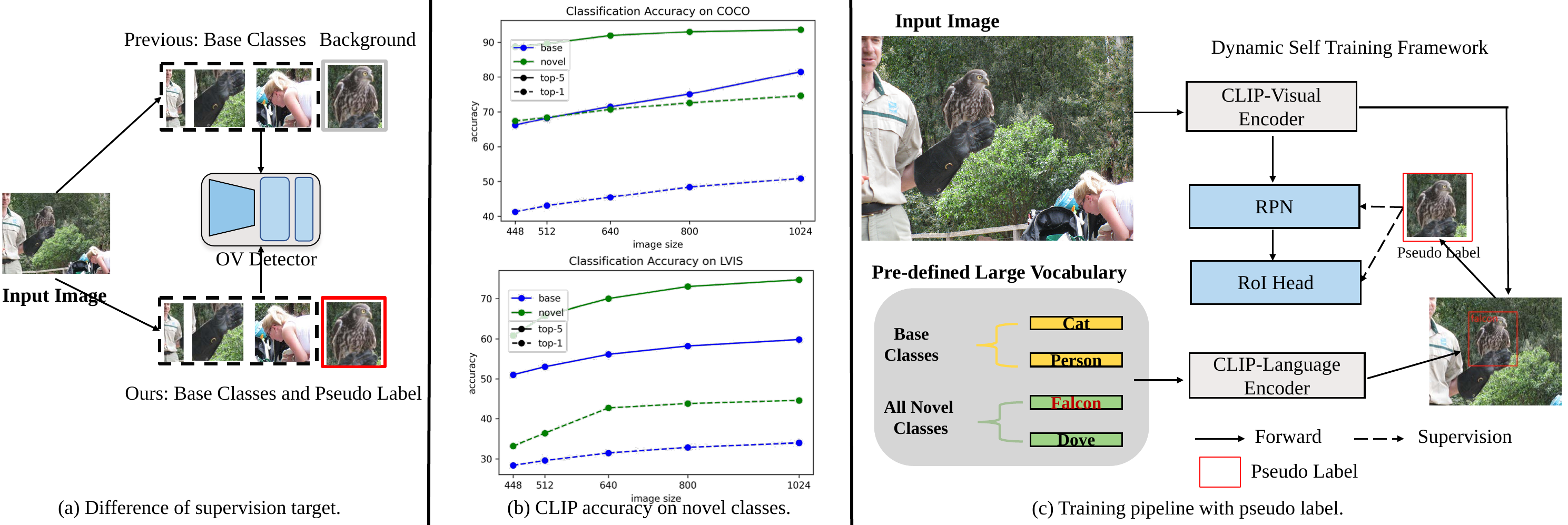}
\caption{Illustration of our motivation and framework. \textbf{(a).} Our DST-Det incorporates novel class labels to supervise the detection head during training. \textbf{(b).} Experiments on OV-COCO and OV-LVIS using CLIP with ground truth box for zero-shot classification. We observe high top-1 and top-5 accuracy in classifying novel classes. \textbf{(c).} Illustration of our dynamic self-training pipeline with the pseudo labels.}
\label{fig:teaser}
\end{figure*}

Object detection is a fundamental task in computer vision, involving localization and recognition of objects within images. 
Previous detection methods~\cite{ren2015faster,maskrcnn,focal_loss} are limited to detecting only predefined categories learned during the training phase. 
This limitation results in a considerably smaller vocabulary compared to human cognition. Although directly extending the categories of large datasets~\cite{lvis_data} would be an ideal solution, it requires an overwhelming amount of human manual annotation. 
Recently, Open-Vocabulary Object Detection (OVOD)~\cite{ovr-cnn,detic,gu2021open_vild,wu2023open} has emerged as a promising research direction to overcome the constraints of a fixed vocabulary and enable the detection of objects beyond predefined categories.

Typical solutions of OVOD rely on pre-trained VLMs~\cite{CLIP, ALIGN}. 
These VLMs have been trained on large-scale image-text pairs and possess strong zero-shot classification capability. 
Prior works~\cite{ovr-cnn, gu2021open_vild, detic} have attempted to leverage VLMs for OVOD by replacing learned classifiers in traditional detectors with text embeddings derived from VLMs. 
Leveraging VLMs' exceptional zero-shot classification ability enables the detector to assign objects to novel classes based on their semantic similarity to the embedded text representations. 
This is the default operation for all OVOD methods.
Moreover, several recent approaches~\cite{gu2021open_vild, VLDet,wu2023baron} aim to distill knowledge from VLMs by aligning individual region embeddings extracted from VLMs via diverse distillation loss designs. 
This alignment strategy transfers zero-shot capability from VLMs to the object detectors. Additionally, some studies~\cite{Kuo2022FVLMOO, xu2022odise,yu2023fcclip, wu2023cora} attempt to build open-vocabulary detectors directly upon frozen visual foundation models. While these methods have shown impressive results in detecting new objects, there is a significant gap between training and testing for novel classes. 
This is because, during training, all novel classes are considered background classes. 
This issue has a particularly crucial effect on datasets with extensive vocabulary sizes. For example, a more recent challenging dataset, V3Det~\cite{wang2023v3det}, contains over 10,000 classes. 
There are over 5,000 classes that are novel classes. However, during training, all these classes are treated as background.

In this work, we rethink the OVOD training pipeline and propose a new dynamic self-training approach by exploiting VLMs' novel class knowledge. 
As shown in Fig.~\ref{fig:teaser}(a), most previous OVOD methods adopt the same training pipeline by using the base class annotations and treating novel objects as background. 
During testing, the object proposals are discovered by the class-agnostic region proposal network (RPN) and then classified using VLMs' zero-shot ability. 
Thus, a conceptual gap exists between the training and testing phases when dealing with novel classes. 
Our approach aims to bridge this gap by classifying all object proposals using a large vocabulary size during both training and testing phases rather than just during testing.
Those proposals are categorized as novel classes that will be treated as pseudo-labels during training.
Since our approach is carried out during the training stage, the pseudo examples are dynamic.
Meanwhile, several previous methods also use pseudo labels. 
As shown in Fig.~\ref{fig:teaser2}(a), previous methods~\cite{vl-plm,pb-ovd} first obtain pseudo labels from unlabeled data or image-caption datasets using VLMs, and then the detector is trained using generated pseudo labels (which contain the novel classes) and base class annotations. 
Our method provides an end-to-end training pipeline and generates pseudo labels dynamically during training. 
We list more features in Tab.~\ref{tab:ps_methods_cmp}.
All methods need both base and novel class names to generate pseudo labels. However, our method is a plug-in-play approach that does not require extra data or trainable components.
In addition, our method also does not require re-training with pseudo-labeled datasets.

\begin{table*}[h]
    \centering
    \caption{Compared with existing methods using pseudo labels for OVD. Our method does not need extra data and extra components. Moreover, our method can be a plug-in method when VLMs are available.}
    \label{tab:ps_methods_cmp}
   \scalebox{0.8}{
    \begin{tabular}{c|c c c c c c}  
    \toprule[0.15em]
     Method &  Extra Data & Extra Components &  Plug-In Method &  Dynamic Labeling & Need Vocabulary Names & Re-training \\
     \hline
     VL-PLM~\cite{vl-plm} & \checkmark  &  \checkmark &   \ding{55} &   \ding{55} & \checkmark & \checkmark  \\
     PB-OVD~\cite{pb-ovd} & \checkmark  &   \ding{55} &  \ding{55} &   \ding{55} & \checkmark  & \checkmark \\
     SAS-Det~\cite{zhao2023improving} &  \ding{55}  & \checkmark & \ding{55} &  \ding{55} & \checkmark & \checkmark  \\
     \hline
     DST-Det (ours) & \ding{55} &  \ding{55} & \checkmark & \checkmark & \checkmark &  \ding{55} \\
    \bottomrule[0.1em]
    \end{tabular}
    }
\end{table*}

%
To substantiate our motivation, we first conduct a toy experiment as illustrated in Fig.~\ref{fig:teaser}(b), where we calculate the top-1 and top-5 accuracy for both base classes and novel classes on LVIS and COCO datasets by using CLIP's visual features and text embeddings. 
Specifically, we utilize the ground-truth boxes to obtain visual features from the feature maps of CLIP and calculate cosine similarity with text embeddings for zero-shot classification. 
Our results show that the top-5 accuracy for novel classes on LVIS and COCO datasets is already \textit{remarkably} high enough to generate credible pseudo labels. This observation inspires us to consider directly using CLIP outputs as pseudo labels during the training phase. 
As shown in Fig.~\ref{fig:teaser}(a, c), we present the DST-Det (dynamic self-training detection) framework, which directly adopts the large vocabulary information provided by CLIP for training. 
The toy experiments also show the potential of our approach if there will be strong VLMs~\cite{open_clip} in the future.

To reduce the extra computation cost of the CLIP visual model, we let the object detection and the pseudo-labeling generation share the frozen CLIP visual backbone. 
This decision was supported by recent developments in exploiting frozen foundation models~\cite{Kuo2022FVLMOO, wu2023cora, yu2023fcclip, OMGSeg}. 
Our DST-Det is based on the two-stage detector Mask R-CNN~\cite{maskrcnn} and incorporates a dynamic pseudo-labeling module that discovers novel classes from negative proposals during training. 
Those proposals with high novel class scores can be considered foreground objects. 
By adopting simple threshold and scoring operations, we can effectively change some background proposals to the foreground according to the class scores. 
These simple designs do not incur any additional learnable parameters for novel class discovery, and the process is dynamic since region proposals vary in each iteration. 
In particular, we apply this operation to both the RPN and the Region-of-Interest Head (RoIhead) in the training stage: during RPN training, we force the discovered novel objects to be foreground objects; during RoIHead training, we add the pseudo labels directly to the classification target. 

We show the effectiveness of DST-Det on the OV-LVIS~\cite{lvis_data}, OV-V3Det~\cite{wang2023v3det} and OV-COCO~\cite{coco_dataset} benchmarks. 
Our proposed method consistently outperforms existing state-of-the-art methods~\cite{VLDet,wu2023baron,Kuo2022FVLMOO} on LVIS and V3Det without introducing extra parameters and inference costs.
Specifically, DST-Det achieves 34.5\% rare mask AP for novel categories on OV-LVIS. With the Faster-RCNN framework~\cite{ren2015faster}, our method achieves 13.5\% novel classes mAP on V3Det, boosting previous methods by 6.8 \% mAP. 
Moreover, our method improves considerably on the smaller vocabulary dataset COCO compared to the baseline method. 
In particular, with recent work~\cite{wu2023clipself} as a strong baseline, our method achieves the new state-of-the-art result with 46.7\% AP on novel classes.
We also provide detailed ablation studies and visual analyses, both of which validate the effectiveness of the DST-Det framework. 
Moreover, we further compare our method with two offline pseudo label methods~\cite{pb-ovd, vl-plm} under the fair comparison setting, where our method still achieves stronger results.

%

%% file: latex/2_related_work.tex
\section{Related Work} 
\label{sec:work}


\noindent
\textbf{Close-Set Object Detection.} Modern object detection methods can be broadly categorized into one-stage and two-stage. One-stage detectors ~\cite{Redmon2015YouOL, Liu2015SSDSS, tian2021fcos, focal_loss, zhou2019objects, zhou2021probablistic, atss, tan2020efficientdet} directly classify and regress bounding boxes using a set of predefined anchors, where anchors can be defined as corners or center points. Recently, several works have adopted query-based approaches~\cite{detr,zhu2020deformabledetr,peize2020sparse,dab-detr} to replace the anchor design in previous works. The query-based approaches use a transformer decoder to decode each object directly without introducing hand-craft anchor boxes. Meanwhile, long-tail object detection aims to address class imbalance issues in object detection. To tackle this challenge, various methods have been proposed, such as data re-sampling~\cite{lvis_data,liu2020deep,wu2020forest}, loss re-weighting~\cite{ren2020balanced,tan2020equalizationv1,tan2021equalizationv2,zhang2021distribution,wang2021adaptive}, and decoupled training~\cite{li2020overcoming,wang2020devil}. However, all these methods cannot be generalized to novel categories. In the real world, we often need a detector to detect and recognize any objects under the large vocabulary scope rather than specific datasets. Our method focuses on the open-vocabulary detection setting.

\noindent
\textbf{Open-Vocabulary Object Detection (OVOD).} This task extends the detector's vocabulary at inference time, where the detector can recognize objects not encountered during the training. Recently, OVOD~\cite{ovr-cnn, gu2021open_vild, zhong2022regionclip, OV-DETR, detic, DetPro, promtdet, wu2023baron, Hanoona2022Bridging, glip, gridclip, pb-ovd, vl-plm, hierkd,Minderer2023ScalingOO,Arandjelovic2023ThreeWT,Zhang2023ASF,Kaul2023MultiModalCF,Cho2023OpenVocabularyOD,Song2023PromptGuidedTF,shi2023edadet,Bravo2022LocalizedVM,Minderer2022SimpleOO,Chen2022OpenVO,Wang2022LearningTD,Buettner2023EnhancingTR,Shi2023OpenVocabularyOD,wu2023clipself,zhou2023rethinking} has drawn increasing attention due to the emergence of vision-language models~\cite{CLIP, ALIGN}. On the one hand, several works~\cite{gu2021open_vild, wu2023baron, DetPro, OV-DETR} attempt to improve OVOD performance by letting the object detector learn knowledge from advanced VLMs. 
For instance, ViLD~\cite{gu2021open_vild} effectively distills the knowledge of CLIP into a Mask R-CNN~\cite{maskrcnn}. 
Then, several following works adopt better knowledge distillation designs.
DetPro~\cite{DetPro} improves the distillation-based method using learnable category prompts, while BARON~\cite{wu2023baron} proposes to lift the distillation from single regions to a bag of regions.
Meanwhile, F-VLM~\cite{Kuo2022FVLMOO} directly builds a two-stage detector upon frozen VLMs (CLIP) and trains the detector heads only. 
It fuses the predicted score and the CLIP score for open-vocabulary recognition.
Moreover, several works~\cite{kim2023contrastive,kim2023region} aim for better VLMs pre-training for OVOD.
Both types of open-vocabulary object detectors treat novel objects as background during training but target them as foreground objects in testing. 
This gap hinders the detector from discovering objects of novel categories. 
We aim to bridge this gap by leveraging stronger pre-trained vision-language models, enabling improved detection performance for novel class detection without extra data or an extra re-training pipeline.

\noindent
\textbf{Vision-Language Pre-training and Alignment.} Several works~\cite{CLIP, ALIGN,kim2021vilt,li2021align,li2022blip} study the pre-training of VLMs in cases of improving various downstream tasks, including recognition, captioning, and text generation. 
In particular, contrastive learning has achieved impressively aligned image and text representations by training VLMs on large-scale datasets, as demonstrated by several works~\cite{CLIP,ALIGN,alayrac2022flamingo}. For example, the representative work, CLIP~\cite{CLIP}, can perform zero-shot classification on ImageNet~\cite{imagenet} without fine-tuning or re-training. 
Inspired by CLIP's generalizability in visual recognition, various attempts have been made to adapt CLIP's knowledge to dense prediction models such as image segmentation~\cite{xu2022odise, rao2022denseclip, OpenSeg, zhou2022maskclip} and object detection~\cite{gu2021open_vild, OV-DETR} in the context of open-vocabulary recognition. Meanwhile, several works~\cite{zhou2022maskclip, shi2022proposalclip} use CLIP to extract pseudo labels for dense prediction tasks. 
Our method also uses the knowledge of the pre-trained VLMs.
Rather than using vision language alignment, our method effectively explores CLIP's novel class-finding ability in the case of OVOD.
In particular, CLIP models help discover the object of novel categories to bridge the conceptual gap between training and testing.

\noindent
\textbf{Pseudo Labeling in Detection and Segmentation.} Pseudo-labeling methods~\cite{kirillov2023segment,Zhu_2019_CVPR,chen2020naive} are well-explored in close-set segmentation and detection. 
Utilizing pre-trained models and unlabeled images achieves even better results than the original labeling models.
For example, recent work SAM~\cite{kirillov2023segment} scaled the unlabeled segmentation dataset using 1\% annotated dataset.
However, even with pseudo labels, via huge data training, SAN generated well in various scenes. 
For open-vocabulary detection and segmentation, several works~\cite{pb-ovd, xpm, zhao2023improving,xie2023mosaicfusion} also propose pseudo labels to improve OVD tasks. 
These approaches firstly pre-train a detector or a semgnenter using base class annotations. 
Then, they generate off-line base and novel labels with the aid of VLM. 
Next, the new models are jointly co-training with both base annotations and generated pseudo labels.
However, these works require \textit{extra VLM tuning and large caption datasets} for pre-labeling, which makes the pipeline complex. 
In contrast, our method is more straightforward and elegant, 
\textit{without} extra pipelines or parameter tuning. 
In addition, compared with default pseudo-labeling methods~\cite{pb-ovd,xpm}, shown in the experiment section, our method performs better on various datasets.

%% file: latex/3_method.tex
\section{Method}
\label{sec:method}
\noindent
\textbf{Overview.} 
This section begins by providing essential background information and discussing the baseline approaches employed in OVOD. We also introduce our baseline methods in detail. Subsequently, we introduce our proposed dynamic self-training method to address the identified issue. Finally, we outline our approach's training and inference details, building upon an existing detector architecture~\cite{maskrcnn}.

\subsection{Preliminaries and Baseline}
\label{sec:preliminary}

\noindent
\textbf{Problem Setting.} 
When given an image $\mathrm{I}$, the object detector should output a set of boxes $b_{i}$ and their corresponding class labels $c_{i}$. Here, $b_i$ is a vector of length four representing the coordinates of the bounding box around the object, $c_i$ is a scalar indicating the class label assigned to that object. During the training phase, the OVOD detector can only access the detection labels of base classes $C_{B}$. But it is required to detect objects belonging to the base classes $C_{B}$ and the novel classes $C_{N}$ at test time. The novel objects are unavailable during the training and are always treated as the background.

\noindent
\textbf{Architecture Overview.}
Most previous OVOD methods adopt a two-stage detector. Therefore, we take the Mask R-CNN~\cite{maskrcnn} as an example to demonstrate how it can be adapted to the OVOD task by leveraging the textual information from pre-trained vision-language models. Mask R-CNN consists of two stages: a Region Proposal Network (RPN) and a Region-of-Interest Head (RoIHead). The RPN denoted as $\mathrm{\Phi}_{\mathrm{RPN}}$ generates object proposals, and the RoIHead denoted as $\mathrm{\Phi}_{\mathrm{RoI}}$ refines the proposals' locations and predicts their corresponding class labels. This process can be formulated as follows: 
\begin{align}
    \label{equ:mask_rcnn}
    & \{r_i\}_{i=1}^M  = \mathrm{\Phi}_{\mathrm{RoI}} \circ \mathrm{\Phi}_{\mathrm{RPN}} \circ \mathrm{\Phi}_{\mathrm{Enc}}(\text{I}), \\ 
    & \{b_i, c_i\}_{i=1}^M  = \{ \mathrm{\Phi}_{\mathrm{box}}(r_i), \mathrm{\Phi}_{\mathrm{cls}}(r_i) \}_{i=1}^M,
\end{align}
where $\mathrm{\Phi}_{\mathrm{Enc}}$ is an image encoder that maps the input image $\text{I}$ to a series of multi-scale feature maps. $M$ is the number of proposals generated by $\mathrm{\Phi}_{\mathrm{RPN}}$. $\mathrm{\Phi}_{\mathrm{RoI}}$ extracts the region embedding $r_i$ from the feature maps given a box proposal. We ignore the Feature Pyramid Network (FPN)~\cite{fpn} for brevity in this paper. Then the box regression network $\mathrm{\Phi}_{\mathrm{box}}$ refines the coordinates of the bounding boxes, and the classification network $\mathrm{\Phi}_{\mathrm{cls}}$ predicts the class label of the object within the bounding box.  We use $\circ$ to represent the cascade of different components.

The classification head in the close-set object detection is learnable in closed-vocabulary object detection and maps the region embedding into predefined classes. 
However, in the open-vocabulary scenario~\cite{detic,gu2021open_vild}, the classifier is substituted with text embeddings generated by pre-trained VLMs and is frozen during the training. The text embedding $t_c$ for the $c$-th object category is generated by sending the category name into a CLIP text encoder using either a single template prompt, "a photo of {category}," or multiple template prompts. And for a region embedding $r$, its classification score of $c$-th category is calculated as follows:
\begin{equation}
\label{equ:cls_score}
    p_c = \frac{\text{exp}(\tau \cdot <r, t_c>)}{\sum_{i=0}^{C} \text{exp}(\tau \cdot <r, t_i>)}
\end{equation}
where $<\cdot,\cdot>$ is the cosine similarity, and $\tau$ is a learnable or fixed temperature to re-scale the value.

\begin{figure*}[t!]
\centering
\includegraphics[width=1.0\textwidth]{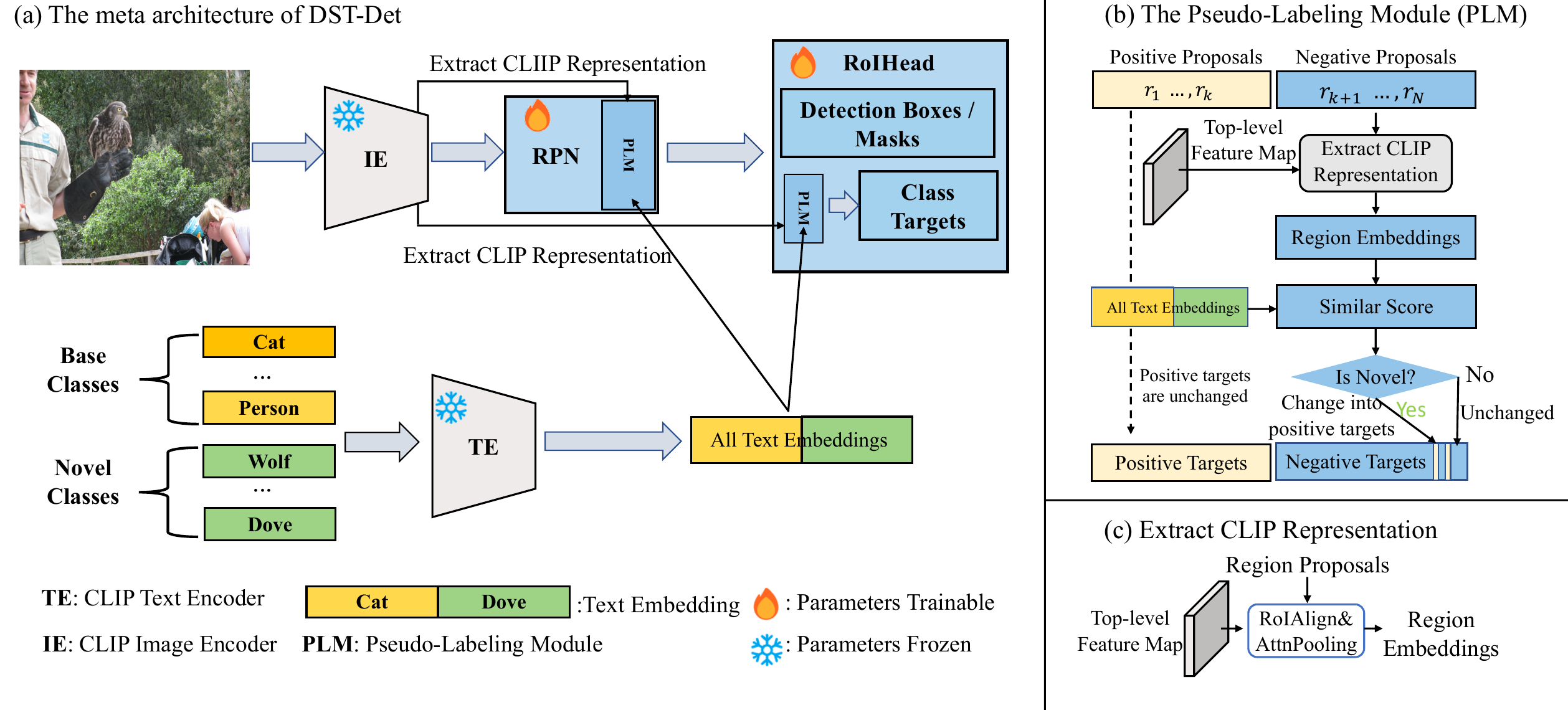}
\caption{Illustration of DST framework. \textbf{(a)} The meta-architecture of DST-Det includes the proposed pseudo-labeling module (PLM), which is integrated into two stages of the detector: RPN and RoIHead. \textbf{(b)} The proposed pseudo-labeling module (PLM). During training, PLM takes the top-level feature map from the image encoder and text embedding of object classes as input and generates the pseudo labels for the RPN and the RoIHead. \textbf{(c)} The process of extracting CLIP representation for region proposals. The RoIAlign operation is applied to the top-level feature map, the output of which is then pooled by the Attention Pooling layer (AttnPooling) of the CLIP image encoder. }
\label{fig:method}
\end{figure*}

\noindent
\textbf{Frozen CLIP as Backbone.}
\label{sec:clip_backbone}
To reduce computation cost and strengthen the open-vocabulary ability of the object detector, we use the CLIP image encoder as the detector backbone in $\mathrm{\Phi}_{\mathrm{Enc}}$. We keep the parameters of our backbone fixed to preserve the vision-language alignment during training following F-VLM~\cite{Kuo2022FVLMOO}. Then, the inference stage of the detector can benefit from both the detection score described in Eq.~\ref{equ:cls_score} and the CLIP score obtained by replacing the region embedding in Eq.~\ref{equ:cls_score} with the CLIP representation of the corresponding region proposal as depicted in Fig.~\ref{fig:method}(c). Specifically, we apply the RoIAlign function $\mathrm{\Phi}_{\mathrm{RoI}}$ to the top-level feature map of $\mathrm{\Phi}_{\mathrm{Enc}}$. Then, the CLIP representations of the region proposals are obtained by pooling the outputs of the RoIAlign via the attention pooling module of the CLIP image encoder. Given a region proposal during the inference, the score of the $c$-th candidate category is obtained by merging the two types of scores via geometric mean:

\begin{equation}
\label{equ:score_fusion}
  s_c = 
    \begin{cases}
        p_c^{1-\alpha}\cdot q_c^\alpha \quad \text{if } i \in C_B \\
        p_c^{1-\beta}\cdot q_c^\beta \quad \text{if } i \in C_N
    \end{cases}
\end{equation}
where $q_c$ is the CLIP score, $\alpha, \beta \in [0,1]$ control the weights of the CLIP scores for base and novel classes, and $C_B$ and $C_N$ represent the base classes and novel classes, respectively.


\subsection{DST-Det: Dynamic Self-Training For OVOD}

\noindent
\textbf{Motivation of DST-Det.}
Previous works have addressed the problem of open-ended classification in OVOD by utilizing the text embeddings from pre-trained VLMs (e.g., CLIP). However, a conceptual gap exists between training and testing for novel classes. As shown in Fig.~\ref{fig:teaser}, annotations for novel objects are considered background during training. However, during the test phase, they are expected to be detected as foreground and classified into a specific novel class based on their detection score.
A new training strategy is needed to bridge the gap between the training and test phases by utilizing CLIP's outputs as pseudo labels for novel classes. 
In the toy experiments depicted in Fig.~\ref{fig:teaser}(b), we treat the ground-truth bounding boxes as region proposals to get the CLIP score for each region proposal and use the CLIP scores to verify the zero-shot classification ability of CLIP on OV-COCO and OV-LVIS datasets. How to get the CLIP score has been illustrated in Sec.~\ref{sec:clip_backbone} and Fig.~\ref{fig:method}(c). 
The results indicate that the top-5 accuracy suffices for discovering novel classes.
This motivates us to consider using CLIP outputs as pseudo labels during training, given the RPN proposals and a large pre-defined vocabulary.

\noindent
\textbf{Pseudo-Labeling Module (PLM).} As shown in Fig.~\ref{fig:method} (b), we categorize proposals into two types: \textit{positive} and \textit{negative}. Positive proposals refer to those having large overlaps with ground truth object bounding boxes and are used to learn box regression and classification, which are supervised by base class annotations.
On the other hand, the negative proposals have a low overlap with the ground-truth boxes but potentially contain objects of novel classes. To avoid treating region proposals of potential novel classes as background contents, our PLM selects negative proposals whose region embeddings have high similarity scores with the text embeddings of novel classes as pseudo labels.

We believe that negative proposals, which have a low overlap with the ground truth boxes, may contain objects belonging to novel classes. Therefore, we introduce the Pseudo-Labeling Module (PLM) to avoid treating these proposals as background. As shown in Fig.~\ref{fig:method}(b) and Fig.~\ref{fig:method}(c), we first extract the CLIP representations of the negative proposals. We treat the negative proposals as regions of interest and use the RoIAlign~\cite{maskrcnn} operation to extract region embeddings from the top-level of the backbone features (CLIP backbone). 
These region embeddings are mapped to the representation space of CLIP text embeddings using a fixed attention pooling layer~\cite{CLIP}.
Then, we calculate the cosine similarity scores between the region embeddings and the text embeddings of the object class names. 
After obtaining the similarity scores, we filter out those proposals classified as base classes or background classes and those with CLIP scores lower than a threshold. 
The remaining few proposals can be identified as novel objects, and we randomly select $\mathrm{K}$ proposals as pseudo labels and change the classification target of these proposals during training. 
The targets of the positive proposals remain unchanged in this process.

\noindent
\textbf{Deployment on RPN and RoIHead.}
Due to the dense anchor head, the RPN stage typically produces numerous negative proposals. To reduce the PLM's computation cost, we leverage Non-Maximum Suppression (NMS) to eliminate redundant proposals and limit the number of negative proposals to a fixed amount, such as 1000. 
In contrast, there are fewer negative proposals in the RoIHead stage compared to the RPN stage. As a result, we send all negative proposals to the PLM during the RoIHead stage. The PLM will select some negative proposals as pseudo-labels according to our strategy and change their classification target. During the RPN stage, we convert the classification target of selected negative proposals from negative to positive. For the classification branch of RoIHead, we change the classification target of selected negative proposals from negative to pseudo labels generated by the PLM. Note that we only apply the pseudo labels produced by PLM to classification losses. The box regression and mask losses remain unchanged for negative proposals, as we cannot create a pseudo box or mask just using VLMs. The detailed process is shown in Fig.~\ref{fig:method} (a).

\noindent
\textbf{Discussion.} The proposed PLM is only used during the training without introducing extra parameters and computation costs during the inference. The pseudo-labeling process requires a vocabulary that contains potential novel object classes. We can either obtain the novel classes from the detection datasets following Detic~\cite{detic} or collect a wider range of object classes from external sources, e.g., the classes defined in image classification datasets~\cite{imagenet}. 

Moreover, the PLM training is single-stage and end-to-end. We list the detailed progress in Algo.~\ref{alg:pseudo_algo} table. In particular, given the negative proposal set $P_{N}$ and corresponding score $S$, our algorithm dynamically selects the potential candidates as novel foreground classes. By only changing the classification loss in both RPN and RoIHeads, our method does not bring any extra trainable parameters and computation costs. In addition, our method can be applied to various RPN-based approaches, which makes it easy to transfer.


\begin{algorithm*}
\caption{Dynamic Self-Pseudo-Labeling Training Process}
\label{alg:pseudo_algo}
\hspace*{\algorithmicindent} \textbf{Input}: negative proposal set $P_N$, score $S$ for each  $p_ \in P_N$, top-level features $F$ from image encoder, text embeddings $t$ for all categories, classification label $L_{cls}$ \\
\hspace*{\algorithmicindent} \textbf{Output}: pseudo classification label $L'_{cls}$

\begin{algorithmic}[1]
    \State candidates = []
    \For {$i$, $p$ \textbf{in} enumerate($P_N$)}
    \If{$S_i > 0$}
        \State $r_i \gets \text{RoIAlign}(F, p)$ \color{black}/* region embeddings */
        \State $s_i \gets \text{MatMul}(r_i, t)$ \color{black}/* calculate potential novel classes scores */
        \If{$\text{max}(s_i) > \text{0.8 and argmax}(s_i) \in \text{novel classes}$} \color{black}/* find the novel objects */
            \State $\text{candidates} \leftarrow i$ \color{black}/* put the novel objects in the training candidates. */
        \EndIf      
    \EndIf
    \EndFor
    \State Keep $\text{selected\_idx} \gets \text{randomly select K RoIs from candidates}$ \color{black}/* select the previous candidates */
 
    \State Update $L'_{cls}[\text{selected\_idx}] \gets \text{pseudo novel classes}$ \color{black}/* classification loss */
    \State Use $L'_{cls}[\text{selected\_idx}]$ to train the RPN and RoIHead for the classification branch. \color{black}/* update both RPN and RoIHead for classification. */
    \State Keep the detection box loss and segmentation mask loss (For Mask-RCNN baseline) unchanged. 
\end{algorithmic}
\end{algorithm*}

\subsection{Training and Inference}
The training losses adhere to the default settings of Mask R-CNN~\cite{maskrcnn}, which comprises three primary losses: classification loss, box regression loss, and mask segmentation loss. Specifically, only the classification loss will be changed based on the pseudo labels, while other losses will remain unmodified. It is important to note that the generated labels are utilized for recognition purposes rather than localization. Therefore, the final loss is expressed as:$ L=\lambda_{\text{ps\_cls}}L_{\text{pl\_cls}} + \lambda_{\text{box}}L_{\text{box}} + \lambda_{\text{mask}}L_{\text{mask}}$. $\lambda$ denotes the weight assigned to each loss, and $L_\text{pl\_cls}$ is our modified pseudo label classification loss. Specifically, we set a weight of 0.9 for the background class and 1.0 for all other classes. For inference, we adopt the same score fusion pipeline (Equ.~\ref{equ:score_fusion}) as stated in Sec.~\ref{sec:preliminary}.

%% file: latex/4_exp.tex
\section{Experiments}
\label{sec:exp}

In this section, we will first introduce the experiment settings and baselines on three different benchmarks. Then, we report results on three different datasets. We also include the comparison experiment results. Next, we explore the extensive ablation studies of our DST-Det. Finally, we further present visualization results and analysis.  

\subsection{Experiment Settings}
\label{sec:exp_set_up}

\noindent
\textbf{Datasets and Metrics.} We carry out experiments on three detection datasets, including OV-LVIS~\cite{gu2021open_vild}, OV-COCO~\cite{ovr-cnn}, and recent challenging OV-V3Det~\cite{wang2023v3det}. For LVIS, we adopt the settings proposed by ViLD~\cite{gu2021open_vild} and mainly report the $\text{AP}_r$, representing the AP specifically for rare categories. We report box AP at a threshold of 0.5 for COCO and mean AP at threshold $ 0.5 \sim 0.95$ for V3Det. All the datasets used in this paper are available online. COCO~\footnote{\url{https://cocodataset.org/}}, LVIS~\footnote{\url{https://www.lvisdataset.org/}}, and V3Det~\footnote{\url{https://v3det.openxlab.org.cn/}} can be downloaded from their official website accordingly.

\noindent
\textbf{Implementation Details.} We follow the settings of F-VLM~\cite{Kuo2022FVLMOO} for a fair comparison. We use the Mask R-CNN~\cite{maskrcnn} framework with feature pyramid network~\cite{fpn} as our detector. All the class names are transferred into CLIP text embedding, following~\cite{gu2021open_vild}. For the "background" category, we input the word "background" into the multiple templates prompt and get a fixed text embedding from the CLIP text encoder. We only use the base boxes for training and all class names as of the known vocabulary as the pre-knowledge of PLM. For the final results on LVIS, we train the model for 59 epochs. For ablation studies, we train the model for 14.7 or 7.4 epochs. For the COCO dataset, we follow the previous works~\cite{Kuo2022FVLMOO}. For the V3Det dataset, we adopt the default setting of the origin baselines~\cite{wang2023v3det}. We have released all the source code and models for further studies.

\noindent
\textbf{Dataset and Training Details on OV-LVIS.}
LVIS~\cite{lvis_data} has a large vocabulary and a long-tailed distribution of object instances. The LVIS dataset contains bounding box and instance labels for 1203 classes across 100k images from the COCO dataset. The classes are categorized into three sets based on their occurrence in training images – rare, common, and frequent. And for open vocabulary object detection, we adopt the settings proposed by ViLD~\cite{gu2021open_vild}. In this approach, annotations that belong to common and frequent categories are categorized as base categories. On the other hand, annotations belonging to rare categories are treated as novel categories. We use the SGD optimizer with a learning rate of 0.36 and a weight decay of 1e-4. We train our model for 46.1k iterations with a batch size of 256.

\noindent
\textbf{Dataset and Training Details on OV-COCO.}
We follow the setting of ovr-cnn~\cite{ovr-cnn} and split COCO2017 into 48 base classes and 17 novel classes. We train our model in this setting for 11.25k interactions with a batch size of 64. Other experiment settings are the same as OV-LVIS.

\noindent
\textbf{Dataset and Training Details on OV-V3Det.}
OV-V3Det~\cite{wang2023v3det} is a vast vocabulary visual detection dataset. It contains extremely large categories, which consist of 13,029 categories on real-world images. The train split of V3Det 1,361,181 objects in 184,523 images, the val split has 178,475 objects in 30,000 images, and the test split has 190,144 objects in 30,000 images.  
For the open vocabulary object detection setting, V3Det randomly samples 6,501 categories as base classes $C_{base}$ and the remaining 6,528 categories as the novel classes $C_{novel}$. For training a detector for V3Det, we use the SGD optimizer with a weight decay of 1e-4, a momentum of 0.9, and a learning rate of 0.02. We train our model for 2x with a batch size of 32.


\begin{table}[ht]
\caption{Results on OV-LVIS benchmark}
\label{tab:res_lvis}
\centering
\scalebox{1.0}{
\begin{tabular}{c|c|c}
    \toprule[0.15em]
    Method& Backbone&mAP$_{r}$\\ 
    \hline
    ViLD~\cite{gu2021open_vild} & RN50 & 16.6 \\
    OV-DETR~\cite{OV-DETR} & RN50 & 17.4 \\
    DetPro~\cite{DetPro} & RN50&19.8 \\
    \textit{VL-PLM}~\cite{vl-plm} & RN50 & \textit{17.2} \\
    OC-OVD~\cite{Hanoona2022Bridging} &RN50& 21.1\\
    OADP~\cite{Wang2023ObjectAwareDP} & RN50 & 21.7\\
    RegionCLIP~\cite{zhong2022regionclip} &RN50x4& 22.0 \\
    CORA~\cite{wu2023cora} &RN50x4& 22.2\\
    BARON~\cite{wu2023baron} & RN50 & 22.6\\
    VLDet~\cite{VLDet} &SwinB&26.3 \\
    EdaDet~\cite{shi2023edadet} &  RN50 & 23.7 \\
    MultiModal~\cite{Kaul23} & RN50 &  27.0 \\
    CORA+~\cite{wu2023cora} & RN50x4 & 28.1 \\
    F-VLM~\cite{Kuo2022FVLMOO}& RN50x64& 32.8\\
    Detic~\cite{detic}&SwinB&33.8 \\
    RO-ViT~\cite{kim2023region} & ViT-H/16 & 34.1 \\ 
    CLIPSelf~\cite{wu2023clipself} & ViT-B/16 & 25.3 \\
    \hline
    Baseline & RN50x16 & 26.3 \\
    DST-Det & RN50x16 & 28.4  (2.1 $\uparrow$) \\
    Baseline & RN50x64 & 32.0 \\
    DST-Det & RN50x64 & 34.5 (2.5 $\uparrow$ ) \\
    \hline
    DST-Det+CLIPSelf~\cite{wu2023clipself} & ViT-B/16 & 26.2 (0.9 $\uparrow$)\\
    \bottomrule[0.1em]
\end{tabular}
}
\end{table}

\begin{table}[ht]
\caption{Results on OV-COCO benchmark}
\label{tab:res_coco}
\centering
\scalebox{1.0}{
\begin{tabular}{c|c|c}
    \toprule[0.15em]
    Method& Backbone&AP$_{50}^{\mathrm{novel}}$\\ \hline
    OV-RCNN~\cite{ovr-cnn}&RN50&17.5\\
    RegionCLIP~\cite{zhong2022regionclip} &RN50 & 26.8\\
    ViLD~\cite{gu2021open_vild}  & RN50& 27.6\\
    Detic~\cite{detic}&RN50 & 27.8 \\
    \textit{PB-OVD}~\cite{pb-ovd} & RN50 & \textit{30.8} \\ 
    \textit{VL-PLM}~\cite{vl-plm} & RN50 & \textit{34.4} \\
    F-VLM~\cite{Kuo2022FVLMOO}& RN50 &28.0\\
    OV-DETR~\cite{OV-DETR} & RN50 & 29.4\\
    VLDet~\cite{VLDet} &RN50 & 32.0\\
    RO-ViT~\cite{kim2023region} & ViT-L/16 &  33.0 \\
     RegionCLIP~\cite{zhong2022regionclip} & RN50x4 & 39.3 \\
     CLIPSelf~\cite{wu2023clipself} & ViT-B/16 & 37.6 \\
     CLIPSelf~\cite{wu2023clipself} & ViT-L/14 & 44.3 \\
    \hline
    Baseline & RN50x64 & 27.4 \\
    DST-Det & RN50x64 & 33.8 (6.4 $\uparrow$ ) \\
    DST-Det + CLIPSelf~\cite{wu2023clipself} & ViT-B/16 & 41.3 (4.7 $\uparrow$)\\
    DST-Det + CLIPSelf~\cite{wu2023clipself} & ViT-L/14 & 46.7 (2.4 $\uparrow$) \\
    \bottomrule[0.1em]
\end{tabular}
}
\end{table}

\begin{table}[th]
    \centering
    \caption{Results on OV-V3Det benchmark}
    \label{tab:res_v3det}
    \begin{tabular}{c|c|c}
        \toprule[0.15em]
        Method& Backbone&AP$^{\mathrm{novel}}$\\ \hline
        Detic~\cite{detic}& RN50 & 6.7 \\
        RegionClip~\cite{zhong2022regionclip} & RN50 & 3.1 \\
        \hline
        Baseline & RN50 & 3.9 \\
        DST-Det & RN50 & 7.2  (3.3 $\uparrow$) \\
        Baseline & RN50x64 &  7.0 \\
        DST-Det &  RN50x64 &  13.5  (6.5 $\uparrow$) \\
        \bottomrule[0.1em]
    \end{tabular}
\end{table}


\begin{table}[t]
    \centering
    \caption{Comparison with Previous Pseudo Labeling Methods. We adopt RN50x64 backbone as strong baselines for each dataset.}
    \label{tab:psu_do_label_exp}
    \begin{tabular}{c|c|c|c}
        \toprule[0.15em]
        Method & COCO  & LVIS & V3Det \\  
        \hline
        Baseline & 27.4 & 32.0 & 7.0 \\
        Method~\cite{pb-ovd} & 29.2 & 32.4 & 7.4  \\
        Method~\cite{vl-plm} & 31.2 & 32.5 & 8.1 \\
        \hline
        Ours & 33.8 & 32.5 & 13.5 \\
        \bottomrule[0.1em]
    \end{tabular}
\end{table}

\begin{table*}[h]
\footnotesize
\centering
\caption{Ablation studies and comparative analysis on LVIS 1.0 dataset. $\mathrm{PLM}^{1}$ means PLM in RPN and $\mathrm{PLM}^{2}$ means PLM in RoI head. By default, we add two PLMs. We mainly report $\text{AP}_{r}$ for reference. $\text{AP}_{all}$ is also used for all classes. All methods use the strong RN50x64 backbone on LVIS. For results on COCO, we report results using RN50.} 
\label{tab:ablation}
\begin{subtable}{0.3\textwidth}
    \centering
    \footnotesize
    \caption{Effectiveness of PLM.}
        \begin{tabular}{c c c c}
            \toprule[0.15em]
            Baseline  & $\mathrm{PLM^{1}}$ & $\mathrm{PLM^{2}}$ & $\text{AP}_{r}$ \\
            \toprule[0.15em]
             \checkmark & - & - & 28.3 \\
              \checkmark& \checkmark & - & 30.4 \\  
             \checkmark & - & \checkmark & 29.5 \\  
           \rowcolor{gray!15}  \checkmark  & \checkmark & \checkmark & 31.8 \\  
            \bottomrule[0.1em]
    \end{tabular}
\end{subtable}
\begin{subtable}{0.3\textwidth}
    \centering
    \footnotesize
    \caption{Loss Choices of PLM.}
    \begin{tabular}{c c c} 
        \toprule[0.15em]
        Setting & $\text{AP}_{r}$ & $\text{AP}_{all}$ \\
        \midrule[0.15em]
        Baseline &  28.3 & 30.2 \\
        box loss &  10.3 & 13.6 \\
        class + box loss &  20.3 & 25.2 \\
        \rowcolor{gray!15}  class loss &  31.8 & 31.5 \\
    \bottomrule[0.1em]
\end{tabular}
\end{subtable}
\begin{subtable}{0.3\textwidth}
    \centering
    \footnotesize
    \caption{Training Schedule. e: epoch.}
    \begin{tabular}{c c c} 
       \toprule[0.15em]
        Setting &  Schedule & $\text{AP}_{r}$ \\
        \midrule[0.15em]
        Baseline &  7.4 e & 28.3 \\
        \rowcolor{gray!15}  w DST &  7.4 e & 32.2 \\
        \hline
        Baseline &  14.7 e & 31.2 \\
        \rowcolor{gray!15}   w DST &  14.7 e  & 33.4 \\
        \bottomrule[0.1em]
    \end{tabular}
\end{subtable}
\vspace{4mm}
\begin{subtable}{0.4\textwidth}
    \centering
    \footnotesize
    \caption{More Design Choices of PLM.}
    \begin{tabular}{c c c} 
        \toprule[0.15em]
        CLIP score  &  $\mathrm{K}$ & $\text{AP}_{r}$\\
        \midrule[0.15em]
         0.5 & 4  &  23.3  \\
        \rowcolor{gray!15} 0.8 & 4  &  31.8\\
         0.4 & 4  &  13.2   \\
         0.8 & 15 &  25.6\\
        \bottomrule[0.1em]
    \end{tabular}
\end{subtable}
\begin{subtable}{0.4\textwidth}
    \centering
    \footnotesize
    \caption{Supervision Vocabulary During Training.}
    \begin{tabular}{c c c} 
        \toprule[0.15em]
        Source & $\text{AP}_{r}$ &$\text{AR}_{all}$  \\
        \hline
        base names only & 28.3 & 30.2 \\
        \rowcolor{gray!15}  using LVIS rare & 31.8 & 31.5 \\
        using ImageNet & 31.5 & 31.2 \\
        \bottomrule[0.1em]
    \end{tabular}
\end{subtable}
\end{table*}

\begin{table*}[h]
\centering
\caption{GFLops and Parameter Analysis. Our method \textbf{does not} bring extra flops or parameters during inference.}
\label{tab:gflops}
\begin{tabular}{c|c|c|c|c|c}
\toprule[0.15em]
 & Method & Backbone & GFLops & Learnable Param. & Frozen Param.\\
\hline
\multirow{4}{*}{Train}& \multirow{2}{*}{Baseline} & CLIP RN50 & 343G &  22.9M & 38.3M \\
& & CLIP RN50x64 &  1565G & 23.9M & 420M \\
\cline{2-6}
& \multirow{2}{*}{Ours} & CLIP RN50 & 561G &  22.9M & 38.3M \\
&  & CLIP RN50x64 &  4961G & 23.9M & 420M \\
\hline
\multirow{4}{*}{Inference} & \multirow{2}{*}{Baseline} & CLIP RN50  & 813G &  22.9M & 38.3M \\
& & CLIP RN50x64 & 8157G & 23.9M & 420M \\
\cline{2-6}
 & \multirow{2}{*}{Ours} & CLIP RN50  & 813G &  22.9M & 38.3M \\
& & CLIP RN50x64 & 8157G & 23.9M & 420M \\
\bottomrule[0.1em]
\end{tabular}
\end{table*}

\subsection{Main Results}
\label{sec:experiment_results}

This section mainly reports the results of three open-vocabulary detection datasets (See Tab.~\ref{tab:res_coco}, Tab.~\ref{tab:res_lvis}, and Tab.~\ref{tab:res_v3det}). Considering that the training data are different, we will further include a fair comparison with the previous pseudo-label methods (See Tab.~\ref{tab:psu_do_label_exp}).

\noindent
\textbf{Results on LVIS OVOD.} Tab.~\ref{tab:res_lvis} shows the results of our approaches with other state-of-the-art approaches on the LVIS dataset. Due to the limited computing power, we use a smaller batch size (64 vs. 256) and shorter training schedule (59 epochs vs. 118 epochs) to build our baseline, which results in 32.0 \% mAP, which is lower than F-VLM~\cite{Kuo2022FVLMOO}. After applying our methods, we obtain 34.5\% mAP on rare classes, about 1.7\%  mAP higher than F-VLM~\cite{Kuo2022FVLMOO}, without introducing any new parameters and cost during inference. Compared with other stronger baselines, including VLDet~\cite{VLDet} and Detic~\cite{detic}, our method does not use any extra data and achieves about 0.7-8.2\% mAP gains. We also find consistent gains over different VLM baselines. We further show the generalization ability of our methods on recent work, CLIPSelf~\cite{wu2023clipself} in the last row of Tab.~\ref{tab:res_lvis}. Without any further introduced parameters and GFLops, our method still improves CLIPSelf by 0.6\% mAP gains. In addition, our method also outperforms previous pseudo method VL-PLM~\cite{vl-plm} by a large marign.

\noindent
\textbf{Results on COCO OVOD.} Tab.~\ref{tab:res_coco} shows the results of our approaches with other state-of-the-art approaches on the COCO dataset. We achieve 27.4\% novel box AP when adopting the frozen Mask R-CNN baseline. After using our offline refinement for the self-training module, our method can achieve a 6.4\% improvement on novel box AP. Both experiment results indicate the effectiveness and generalization ability of our approach. Moreover, we also verify our method on a stronger detector, CLIPSelf~\cite{wu2023clipself}. 
With a stronger VLM as the backbone and a stronger detector as the baseline, our method still achieves 2.4-4.7\% improvement on novel classes, which results in a new state-of-the-art result using \textbf{only} COCO datasets.
This indicates our proposed method is a general and plug-in component to improve VLM-based OVOD methods.

\noindent
\textbf{Results on V3Det OVOD.} V3Det is a more challenging dataset with a larger vocabulary size than both LVIS and COCO. As shown in Tab.~\ref{tab:res_v3det}, our methods achieve new state-of-the-art results on different backbones. In particular, with RN50x64, our method achieves 13.5 \% mAP on novel classes, significantly outperforming previous STOA by 6.8 \%. Moreover, with different backbones, our methods can still improve the strong baseline via 3.3-6.5\%.

\noindent
\textbf{Comparison to Previous Pseudo Labeling Method.} Although in Tab.~\ref{tab:res_coco} and Tab.~\ref{tab:res_lvis} our method outperform previous pseudo labeling methods, both the training data and baseline are different. Thus, in Tab.~\ref{tab:psu_do_label_exp}, we compare the previous method under a fair setting, including using the same detector in the same codebase. For the table, compared with previous pseudo labeling methods~\cite{pb-ovd,vl-plm}, we find our approach works best, even without re-training. Moreover, we find our approach is more effective on larger vocabulary datasets, including LVIS and V3Det. This is because, in those datasets, rare classes are treated as background during the training. Thus, a simple trained detector cannot generalize well without extra text-image datasets.

\begin{figure}[t]
\centering
\includegraphics[width=0.50\textwidth]{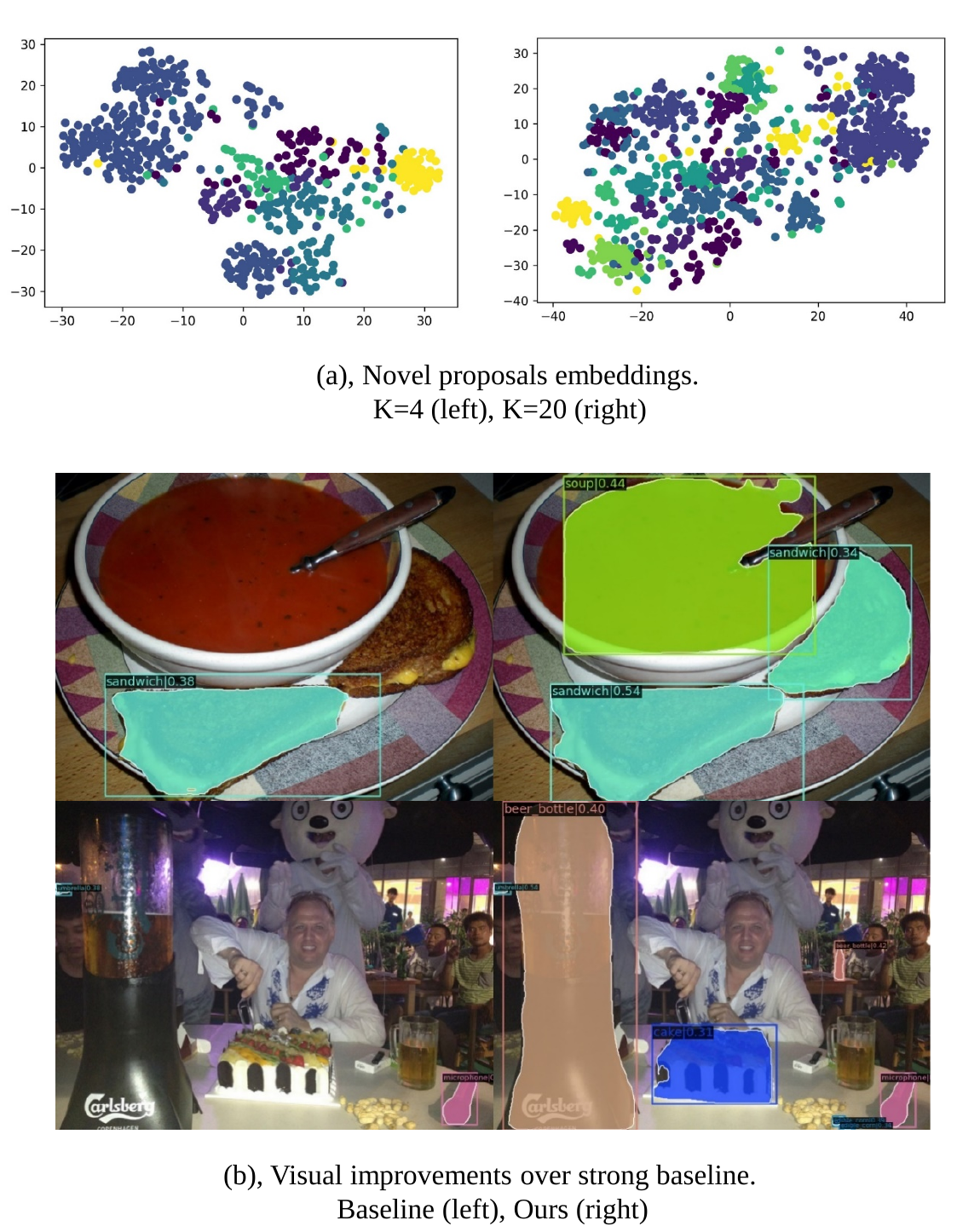}
\caption{Visual Analysis of DST framework. (a), We present a t-SNE analysis on the novel region embeddings during training.  Different colors represent different classes. We find that using fewer training samples works well. (b), We show visual improvements over the strong baseline. Our method can detect and segment novel classes, as shown on the right side of the data pair.}
\label{fig:visual_exp_results}
\end{figure}

\begin{figure*}[t!]
\centering
\includegraphics[width=0.90\textwidth]{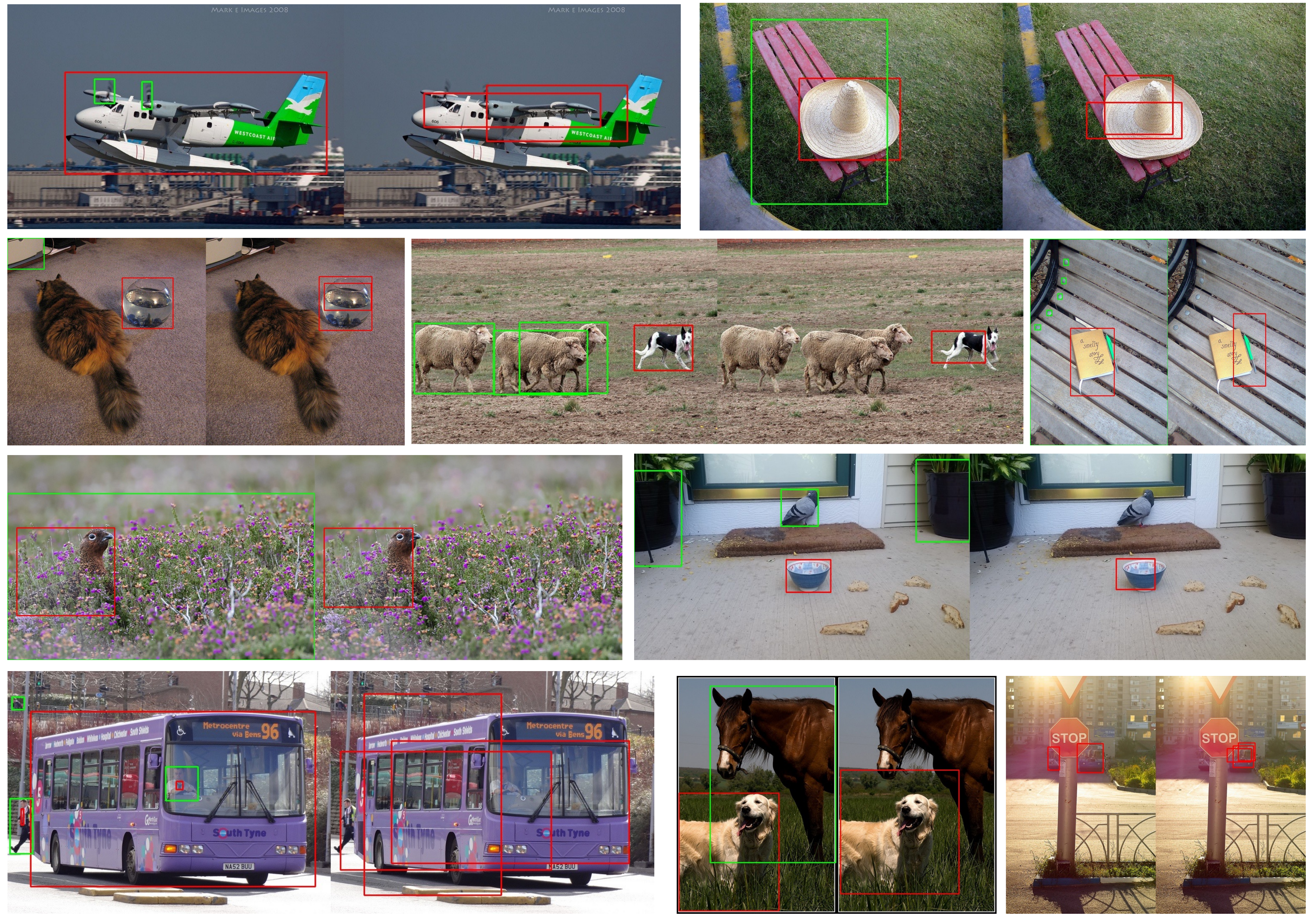}
\caption{Pseudo Label Visual Examples. Left: We visualize the class-agnostic ground truth bounding boxes. The green boxes represent the ground truth of base classes and will be used as foreground supervision, while the red boxes represent the ground truth of possible novel classes that are not allowed during training. Right: The red boxes represent the pseudo labels we selected from negative proposals.}
\label{fig:plm_vis}
\end{figure*}

\subsection{Ablation Study and Analysis}
\label{sec:ablation_studies}

In this section, we conduct detailed ablation studies and analyses on our DST-Det.

\noindent
\textbf{Effectiveness of PLM.} In Tab.~\ref{tab:ablation}(a), we first verify the effectiveness of PLM. Adding PLM in RPN obtains 1.9 \% mAP improvements while inserting PLM in RoI heads leads to 1.2\% gains. This indicates the novel classes are more sensitive to RPN. As shown in the last row, combining both yields better results, confirming the orthogonal effect of the two classification losses. As a result, the PLM results in a notable improvement of over 3.0\%.

\noindent
\textbf{Ablation on Loss Design in PLMs.} In PLM, we only adopt the classification loss of generated labels and ignore the box loss. This is motivated by the fact that most generated proposals are inaccurate, adversely affecting the localization ability developed using normal base annotations. To validate this assertion, we add the box loss into the PLM. As shown in Tab.~\ref{tab:ablation}(b), we observe a significant drop in both $\text{AP}_{r}$ and $\text{AP}_{all}$. Furthermore, our objective is to enable the detector to recognize the novel objects or part of the novel objects. As shown in Fig.~\ref{fig:visual_exp_results}, repetitive proposals still persist even with filtering operations in PLMs.

\noindent
\textbf{Ablation on Various Schedules.} In Tab.~\ref{tab:ablation}(c), we verify the effectiveness of longer schedule training. After extending the schedule into 14.7 epochs, we still observe over 2.0\% improvements. Finally, we extend the training schedule to 59 epochs and still obtain 2.5\% improvements, as shown in Tab.~\ref{tab:res_lvis}. This indicates our method can be scaled up with longer training and stronger baselines.

\noindent
\textbf{Ablation on Effect of Pseudo Score and Pseudo Label Number.} We adopt CLIP score and training samples $\mathrm{K}$ to select the possible proposals, since most are background noises. As shown in Tab.~\ref{tab:ablation}(d), decreasing the CLIP score and increasing training samples lead to inferior performance. Decreasing the score may involve more irrelevant objects or context. Increasing training examples may result in more repeated and occluded objects. Moreover, we also visualize the embeddings of novel proposals during the training with different $\mathrm{K}$ in Fig.~\ref{fig:visual_exp_results}(a). With more training examples, the novel class embeddings become less discriminative, contributing to increased noise. Thus, we set $\mathrm{K}$ to 4 and the CLIP score to 0.8 by default.

\noindent
\textbf{Ablation on Source of Vocabulary.} We also verify the vocabulary source for novel classes in Tab.~\ref{tab:ablation}(e). We experiment with two types of sources: the classes (including rare classes) defined in LVIS~\cite{lvis_data} and the classes defined in ImageNet~\cite{imagenet}. As shown in Tab.~\ref{tab:ablation}(e), our approach achieves considerable performance even when we \textit{do not} have access to the novel classes in the test dataset and use an external source of potential novel classes, i.e., ImageNet. The source of large vocabulary maybe more important when adopting multi-dataset co-training.


\noindent
\textbf{GFLops and Parameter Analysis.}
In Tab.~\ref{tab:gflops}, we list the GFLops and number of parameters during training and inference. Our method uses a frozen backbone and a learnable detection head, and the parameters mainly come from the backbone. When using a large backbone, CLIP RN50x64~\cite{CLIP}, the number of learnable parameters accounts for only one-twentieth of the total number. During training, our method will get 512 proposals per image and obtain its embeddings through RoIAlign~\cite{maskrcnn} and \textit{AttentionPool} operation of CLIP. Compared to the baseline model, our method contributes a significant amount of GFLops from \textit{AttentionPool} operation. During inference, we use the same inference pipeline in the baseline method and use 1000 proposals per image. A larger number of proposals leads to higher GFLops.


\subsection{Visualization, Error Analysis, and Future Work}

This section mainly presents visual examples of visual improvements, pseudo-label examples, visual analysis, and failure case analysis. In addition, we also point out the impact of our work, limitations, and future work.

\begin{figure*}
    \centering
    \includegraphics[width=0.80\textwidth]{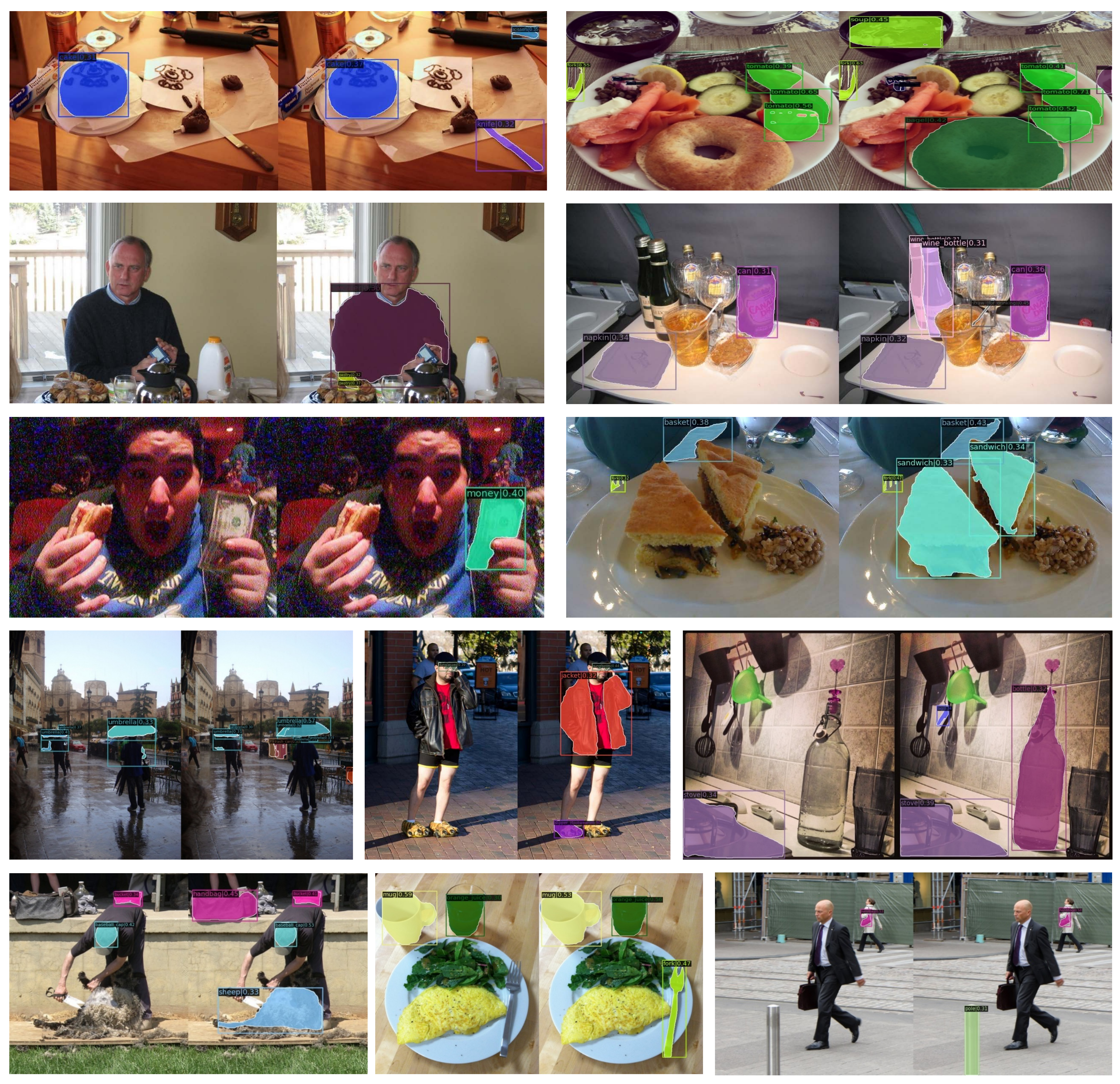}
    \caption{Improvement examples compared to baseline on LVIS dataset. The left is the baseline, and the right is our method.}
    \label{fig:supp_improvement}
\end{figure*}

\noindent
\textbf{Visual Improvements Analysis.} In Fig.~\ref{fig:visual_exp_results} (b), we present several visual improvements over a strong baseline (32.0 $\text{AP}_{r}$) on LVIS. Compared to the baseline, our method demonstrates effective detection and segmentation of novel classes, such as soup and beer bottles. In Fig.~\ref{fig:supp_improvement}, we present more visual examples of the LVIS datasets. These classes are rare and do not have any annotations during the training, which proves the effectiveness of finding novel and rare classes.

\begin{figure*}[h!]
\centering
\includegraphics[width=0.80\textwidth]{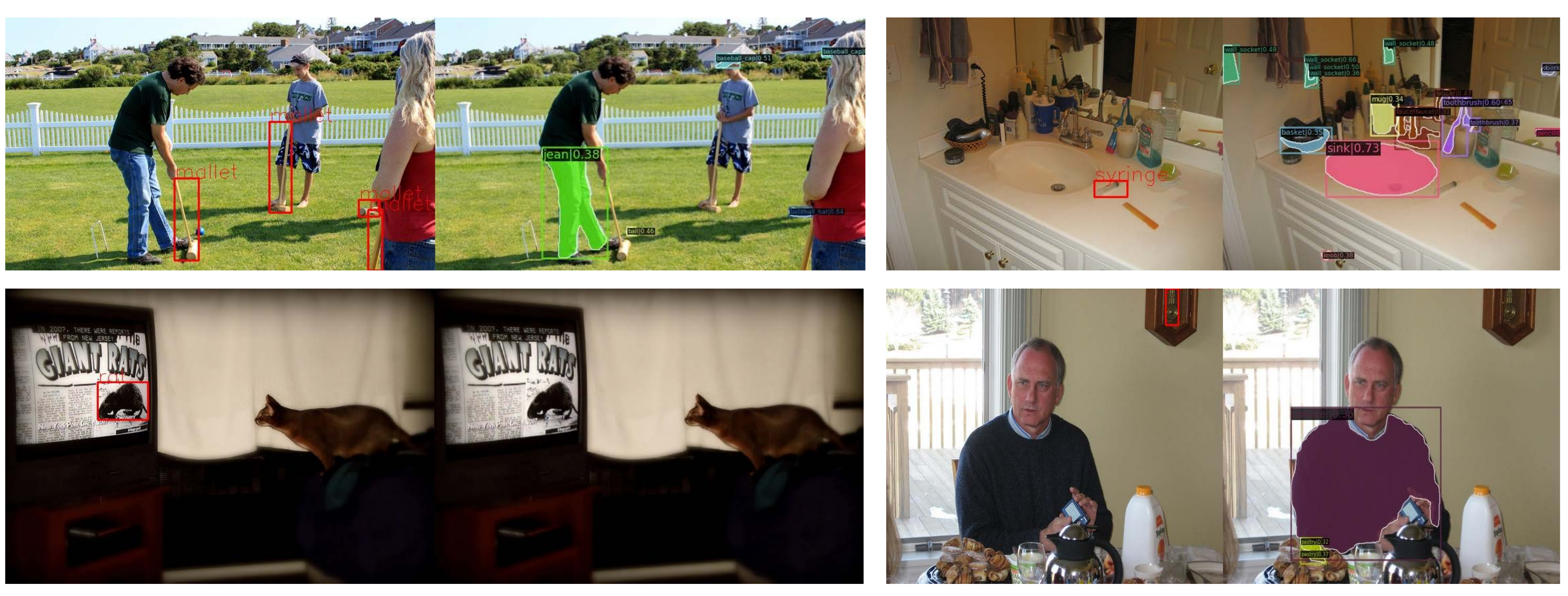}
\caption{Failure case visualization. The red boxes on the left images represent the ground truth annotations of novel classes; the right images are our predictions.}
\label{fig:failure_case}
\end{figure*}


\noindent
\textbf{Qualitative Novel Class Example in DST.} We further visualize the generated novel examples in Fig.~\ref{fig:visual_exp_results}. Most visual examples exhibit satisfactory quality after the filtering operation in PLMs. Compared with previous methods, which treat the novel classes as background, our method directly forces the detector to train the classification head with the assistance of VLM. Despite overlapping or repetitive proposals, we can successfully train the detector by modifying only the classification loss within the RPN and RoI heads. 

\noindent
\textbf{Failure Cases Visualization.}
In Fig.~\ref{fig:failure_case}, we present several failure examples on LVIS. Although our method introduced the pseudo-labeling module for novel classes during training, many novel class objects cannot be detected or classified correctly. We argue that adopting stronger VLMs can solve this issue.

\noindent
\textbf{Pseudo Labels Visualization.} This section presents additional visualization results. Specifically, we present the pseudo labels generated by our pseudo-labeling module in Fig.~\ref{fig:plm_vis}. The visualization results are alongside the ground truth annotations for comparison. Our pseudo labels can successfully recall the novel class annotations not seen during training.

\noindent
\textbf{Limitation and Future Work.} Our proposed approach leverages a better VLM model to generate high-quality pseudo labels, where the zero-shot ability of VLMs affects the quality of novel class labels. However, with more strong VLMs in the future~\cite{CLIP,li2022scaling}, our method is more feasible for large vocabulary applications. Another direction is to explore our framework under multi-dataset detection using one model with a large vocabulary size, which will fully utilize the potential of VLM's vocabulary scope to generate more accurate labels.

\noindent
\textbf{Board Impact.} Our method designs the first dynamic self-training on OVOD via mining the self-knowledge from the VLM model and RPN heads. Without extra training data or learnable components, we obtain significant boosts among strong baseline on COCO and LVIS datasets. Our method can be easily extended into other related domains, including open vocabulary instance/semantic segmentation. Rather than achieving STOA results, our goal is to fully explore the potential of VLM in the detector, which makes our method a generalized approach for various VLM and detectors.

%% file: latex/5_conclusion.tex
\section{Conclusion}
\label{sec:conclusion}
This paper presents DST-Det, a novel open vocabulary object detection method incorporating a dynamic self-training strategy. By rethinking the pipeline of previous two-stage open vocabulary object detection methods, we identified a conceptual gap between training and testing for novel classes. To address this issue, we introduce a dynamic self-training strategy that generates pseudo labels for novel classes and iteratively optimizes the model with the newly discovered pseudo labels. Our experiments demonstrate that this approach can significantly improve the performance of mask average precision for novel classes, making it a promising approach for real-world applications. We hope our dynamic self-training strategy can help the community mitigate the gap between the training and testing for the OVOD task.

\noindent
\textbf{Acknowledgement.} We thank the idea discussion of several friends, including Guangliang Cheng, Yining Li, and Kai Chen, in the earlier drafts. This work is supported by the National Key Research and Development Program of China (No.2023YFC3807603). This study is also supported under the RIE2020 Industry Alignment Fund Industry Collaboration Projects (IAF-ICP) Funding Initiative, as well as cash and in-kind contributions from the industry partner(s). It is also supported by Singapore MOE AcRF Tier 2 (MOE-T2EP20120-0001). We also gratefully acknowledge the support of SenseTime Research for providing the computing resources for this work. 
